\documentclass[letterpaper, 10 pt, journal, twoside]{IEEEtran}

\usepackage{cite}
\usepackage[switch]{lineno} 
\usepackage{algorithm}
\usepackage{algpseudocode}
\usepackage{graphicx} 
\usepackage{times} 
\usepackage{amsmath} 
\usepackage{amssymb}  
\usepackage[labelformat=simple]{subcaption}

\usepackage{tabularray}
\usepackage{tabularx}
\usepackage{soul}
\usepackage{xcolor}

\usepackage[table]{xcolor}
\usepackage{amsmath}

\definecolor{headerblue}{RGB}{31, 102, 128}
\definecolor{rowgrey}{RGB}{218, 222, 228}

\begin{document}

\title{Beyond Self-Play: Hierarchical Reasoning for Continuous Motion in Closed-Loop Traffic Simulation} 

\author{Weifan Zhang\textsuperscript{\dag}, Xiaofeng Zhao\textsuperscript{\dag}, Adel Bazzi, Mingrui Li, Yifan Wei, and Dengfeng Sun
\thanks{\textsuperscript{\dag}Weifan Zhang and Xiaofeng Zhao contributed equally to this work.}%
\thanks{W. Zhang, X. Zhao, A. Bazzi, M. Li, and Y. Wei are with the Isuzu Technical Center of America, Plymouth,
MI, USA (e-mail: Weifan.Zhang@isuzu.com; Xiaofeng.Zhao@isuzu.com; Adel.Bazzi@isuzu.com; Mingrui.Li@isuzu.com; Yifan.Wei@isuzu.com).}%
\thanks{D. Sun are with the School of Aeronautics and Astronautics, Purdue University, West Lafayette, IN 47907, USA (e-mail: dsun@purdue.edu).}%
\thanks{Corresponding author: Xiaofeng Zhao.}%
}

\markboth{IEEE Robotics and Automation Letters. Preprint Version.}%
{Zhang \MakeLowercase{\textit{et al.}}: Beyond Self-Play: Hierarchical Intent and Continuous Motion...}

\maketitle

\begin{abstract}
Closed-loop traffic simulation requires agents that are both scalable and behaviorally realistic. Recent self-play reinforcement learning approaches demonstrate strong scalability, but their equilibrium strategies fail to capture the socially aware behaviors of real human drivers. We propose a hierarchical architecture that goes beyond self-play by combining high-level multi-agent interaction reasoning with low-level continuous trajectory realization. Specifically, a Stackelberg-style Multi-Agent Reinforcement Learning (MARL) module generates interaction-aware intention commands. These commands condition a low-level continuous motion module, translating the strategic intent into physically consistent, scene-responsive control sequences. To mitigate distribution shift in closed-loop deployment, we introduce a hybrid co-training scheme combining MARL with auxiliary recovery supervision. Experiments on a SUMO-based urban network demonstrate that the proposed framework achieves superior control smoothness and safety compared to self-play and passive imitation baselines, while maintaining competitive traffic efficiency.
\end{abstract}


\section{INTRODUCTION}
Naturalistic closed-loop traffic simulation faces fundamental challenges in balancing behavior realism and scalability, particularly in sparse data regions \cite{liu2024curse}. A promising direction toward scalability emerged from recent work \cite{cusumano2025robust}, which demonstrated that a general, single reactive policy trained via large-scale self-play reinforcement learning can outperform state-of-the-art specialists relying on fine-grained reward shaping and heavy scenario engineering. However, task performance and behavioral realism are distinct objectives: self-play converges to equilibrium strategies that need not reflect the hierarchically structured, socially aware behaviors of real human drivers, which resist capture by a flat policy  \cite{cornelisse2025building, wang2026learning}. This highlights a fundamental modeling fidelity problem: the goal is not merely to produce a collision-free agent, but to simulate how a multi-agent scene evolves under different decisions, ensuring that agent interactions remain realistic and consistent. In closed-loop interactive simulation, an agent's decision has long-term consequences through the subsequent reactions of others. Therefore, developing an interaction policy with the appropriate inductive biases to bridge this gap in long-term fidelity remains a critical open challenge.

Recent advances have highlighted the critical role of interaction-aware modeling in closing this gap. In particular, combining imitation learning for behavioral realism with reinforcement learning for scalable joint objective optimization has emerged as a promising direction \cite{lu2023imitation, zhang2023learning, pei2025advancing, booher2024cimrl}. Transformer-based architectures like Wayformer \cite{Nayakanti2022WayformerMF} and MotionLM \cite{seff2023motionlm} have improved spatial-temporal multi-agent prediction, while game-theoretic architectures such as GameFormer \cite{huang2023gameformer} and STEER \cite{zhang2024sequential} more explicitly model inter-agent dependencies and temporally causal rollouts. However, despite these advances, integrating high-level strategic reasoning with continuous trajectory generation typically falls into two constrained paradigms. On one hand, many Multi-Agent Reinforcement Learning (MARL)  approaches rely on discrete action spaces or rule-based motion primitives to maintain tractability \cite{qiao2021behavior, li2025survey, Emergency}. This structurally limits the policy, preventing the generation of flexible, fine-grained behaviors needed in highly interactive scenarios \cite{Cooperation}. On the other hand, continuous imitation learning approaches excel at reproducing realistic short-term trajectories, but they do not explicitly model how an agent's decisions causally affect the future responses of others \cite{lu2023imitation, spencer2021feedback, de2019causal}. Ultimately, both paradigms struggle to capture long-horizon interaction dynamics, where small early deviations propagate into substantially different downstream behaviors \cite{suo2021trafficsim, wu2024smart}.

These observations motivate a hierarchical formulation that separates high-level interaction reasoning from low-level continuous trajectory realization. In this paper, we propose a framework in which high-level strategic reasoning conditions a continuous motion model. Specifically, we use MARL with a Stackelberg-style formulation to model interaction-aware decision-making, and utilize an imitation method to generate trajectories conditioned on high-level strategic commands. In this formulation, the high-level module captures interaction-aware intent and explicitly models agent responses, while the low-level motion model realizes these decisions as physically consistent trajectory predictions that remain responsive to scene context. To improve closed-loop stability, we further employ a hybrid co-training scheme that uses heuristic recovery trajectories as auxiliary supervision for the low-level motion model. This decomposition reduces the burden of direct continuous multi-agent policy learning while maintaining efficient and continuous trajectory generation for closed-loop simulation.

The main contributions of this work are as follows:
\begin{itemize}
    \item We propose a hierarchical framework that combines high-level multi-agent interaction reasoning with low-level continuous trajectory realization for interactive traffic simulation.
    \item We introduce a low-level realization module that conditions continuous trajectory realization on high-level strategic commands, enabling flexible motion generation from discrete decision inputs and demonstrating the necessity of high-level interaction reasoning.
    \item We demonstrate improved long-horizon closed-loop performance in complex multi-agent scenarios compared to passive or implicitly modeled interaction baselines.
\end{itemize}

\section{PRELIMINARIES}
\subsection{Spatio-Temporal Sequential Markov Game}
We adopt the spatio-temporal sequential Markov game (STMG) formulation to model ordered multi-agent decision-making. Let $\mathcal{I}=\{1,\dots,N\}$ denote the set of agents, $s_t\in\mathcal{S}$ the environment state, and $a_t=(a_t^1,\dots,a_t^N)$ the joint action. Given an ordering $H=(h_1,\dots,h_N)$, agent $h_i$ acts conditioned on the current state and the decisions of preceding agents. The corresponding sub-game state is defined as
\[
s_t^{h_i} = (s_t, a_t^{h_1}, \dots, a_t^{h_{i-1}}),
\]
and the joint policy factorizes sequentially as
\[
\pi(a_t \mid s_t)=\prod_{i=1}^{N}\pi^{h_i}(a_t^{h_i}\mid s_t^{h_i}).
\]
This formulation captures directional dependencies among agents through ordered conditioning and is adopted in STEER to support Stackelberg-style autoregressive multi-agent decision-making.

\subsection{Stackelberg Decision Structure}
We further adopt the Stackelberg perspective, in which later decisions are conditioned on preceding ones. In the classical two-agent case, this corresponds to a bi-level optimization where a follower selects its policy in response to the leader's action. More generally, this view provides a principled basis for modeling conditional and asymmetric interactions through sequential decision dependence rather than simultaneous joint action selection.

\subsection{Wayformer}
Wayformer is a transformer-based architecture for multi-agent motion prediction. Given historical agent states $\mathbf{X} \in \mathbb{R}^{N \times T_h \times D_x}$ and scene context $\mathbf{M}$, it encodes spatial-temporal interactions through attention-based fusion and predicts future trajectories over a horizon $T_f$. Formally, Wayformer maps
\[
\mathbf{Z} = f_{\text{enc}}(\mathbf{X}, \mathbf{M}), \qquad
\mathbf{Y} = f_{\text{dec}}(\mathbf{Z}),
\]
where $\mathbf{Z}$ is a latent scene representation and $\mathbf{Y} \in \mathbb{R}^{N \times T_f \times D_y}$ denotes the predicted future trajectories. In this work, Wayformer serves as the continuous motion prediction backbone.

\section{METHODS}
 \begin{figure}[t]
    \centering       \includegraphics[width=.95\linewidth]{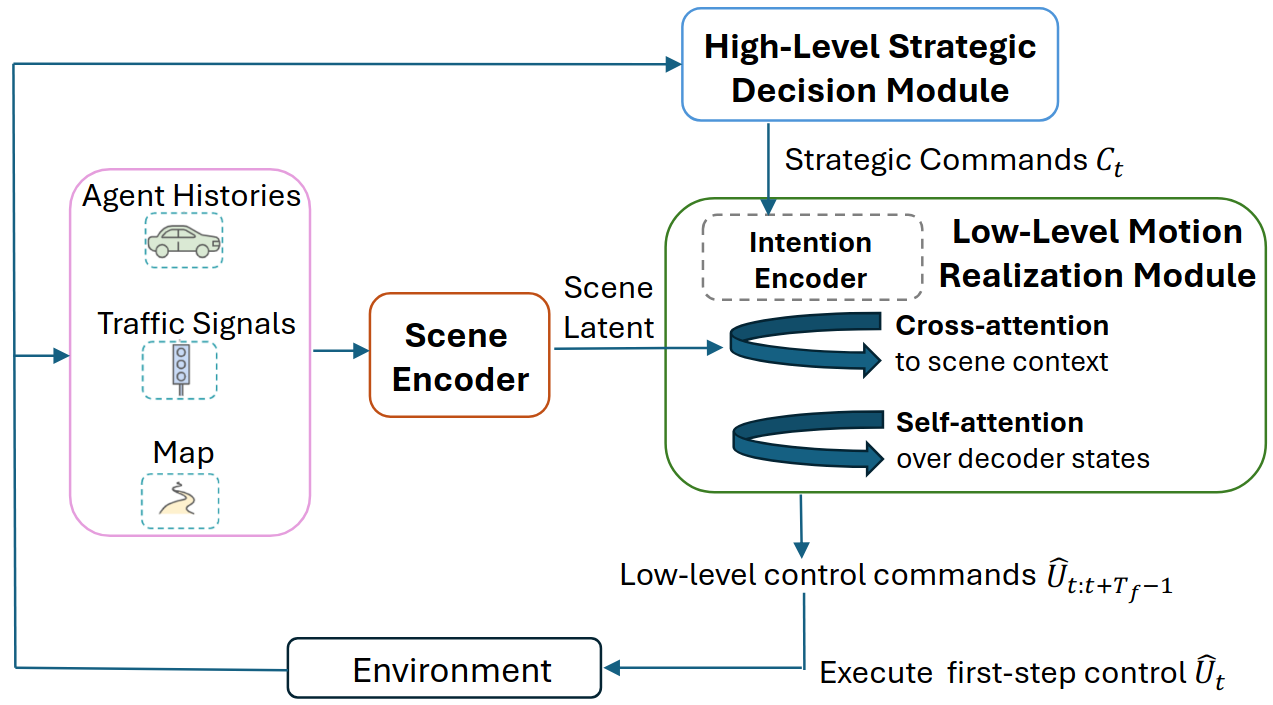}
    \caption{The high-level strategic decision module generates strategic commands $c_t$, which are injected into a command-conditioned low-level motion realization module together with the scene latent. The low-level module predicts a finite-horizon control rollout $\hat{\mathbf{U}}_{t:t+T_f-1}$, of which only the first-step control $\hat{\mathbf{U}}_t$ is executed in a sliding-window closed-loop manner.}
   \label{fig: 31}
 \end{figure}
 
\subsection{High-Level Strategic Decision Module}
Directly learning continuous joint policies for many interacting agents is difficult due to the large joint action space, non-stationarity, and long-horizon coupling among agent decisions. To improve tractability, we introduce a high-level strategic decision layer that operates over an ordered multi-agent interaction structure. Building on the STMG formulation, we define a command-level sub-game state for agent $h_i$ under ordering $H=(h_1,\dots,h_N)$ as
\[
s_t^{h_i} = \bigl(s_t, c_t^{h_1}, \dots, c_t^{h_{i-1}}\bigr),
\]
where $s_t$ is the current environment state and $\{c_t^{h_1},\dots,c_t^{h_{i-1}}\}$ are the strategic commands generated by preceding agents. The high-level policy then produces a strategic command autoregressively:
\[
c_t^{h_i} \sim \pi_{\mathrm{high}}^{h_i}(c \mid s_t^{h_i}).
\]

In our implementation, the command is defined as
\[
c_t^{h_i} = \bigl(a_{RL}^{h_i}, W_{ref}^{h_i}\bigr),
\]
where $a_{RL}^{h_i}$ denotes a discrete high-level maneuver decision and $W_{ref}^{h_i}$ denotes waypoint-level guidance. Intuitively, $a_{RL}^{h_i}$ captures coarse decision semantics such as whether the agent should maintain progress, yield, or switch maneuver mode, while $W_{ref}^{h_i}$ provides a geometric target that refines the intended behavior at a higher abstraction level than direct control.

To model ordered dependency among agents, we implement $\pi_{\mathrm{high}}$ using a Stackelberg-style MARL policy inspired by STEER. Rather than making all agents decide simultaneously, this autoregressive coordination mechanism allows later agents to condition their strategic decisions on preceding commands. Such ordered conditioning reduces strategic ambiguity in dense scenes by explicitly exposing directional inter-agent dependencies at the decision level. The role of this high-level module is therefore not to output final controls, but to produce a compact interaction-aware command interface that is easier to optimize than direct multi-agent continuous control.

\subsection{Command-Conditioned Wayformer}
The low-level module realizes the strategic commands as executable controls while remaining responsive to the surrounding scene. We adopt Wayformer as the continuous motion backbone because it compactly encodes rich spatial-temporal context across many interacting agents and map elements. Let
\[
\mathbf{X}_t \in \mathbb{R}^{N \times T_h \times D_x}
\]
denote the historical kinematic states of all $N$ agents over a history horizon $T_h$, and let $\mathbf{M}_t$ denote the scene context, including map geometry and traffic-signal information. A scene encoder first maps these inputs into a latent scene representation:
\[
\mathbf{Z}_t = f_{\mathrm{enc}}(\mathbf{X}_t,\mathbf{M}_t).
\]
The latent $\mathbf{Z}_t$ summarizes the current multi-agent scene and serves as the context on which low-level control generation is based.

To inject strategic intent, we introduce an intention encoder that transforms the high-level commands
\[
\mathbf{C}_t = \{c_t^1,\dots,c_t^N\}
\]
into intention embeddings. These intention embeddings are then fused with the scene latent through a modified Wayformer decoder. Specifically, the decoder uses cross-attention to condition its internal states on the scene latent and self-attention to refine decoder states under the influence of the injected commands. This produces a command-conditioned finite-horizon control rollout:
\[
\hat{\mathbf{U}}_{t:t+T_f-1} = f_{\mathrm{dec}}(\mathbf{Z}_t,\mathbf{C}_t),
\]
where
\[
\hat{\mathbf{U}}_{t:t+T_f-1}
=
\{\hat{\mathbf{U}}_t,\hat{\mathbf{U}}_{t+1},\dots,\hat{\mathbf{U}}_{t+T_f-1}\},
\]
and each step consists of multi-agent low-level controls (acceleration and steering angle):
\[
\hat{\mathbf{U}}_{\tau}=\{\hat{u}_{\tau}^1,\dots,\hat{u}_{\tau}^N\}, \qquad
\hat{u}_{\tau}^i=(\mathrm{acc}_{\tau}^i,\delta_{\tau}^i).
\]

This design separates \emph{what} should be done from \emph{how} it is physically executed. The high-level module determines an interaction-aware strategic command, while the low-level module realizes that command as scene-consistent control sequences. In contrast to a passive predictor that extrapolates likely motion from history alone, the command-conditioned Wayformer acts as a controllable motion realization module: it preserves the strong scene understanding of Wayformer while making its outputs directly responsive to strategic intervention.

\subsection{Sliding-Window Closed-Loop Execution}
We deploy the framework in a sliding-window closed-loop manner. At each step $t$, the high-level module predicts strategic commands $\mathbf{C}_t$ from the current scene, and the command-conditioned low-level module predicts a finite-horizon control sequence
\[
\hat{\mathbf{U}}_{t:t+T_f-1}
= f_{\mathrm{dec}}(\mathbf{Z}_t,\mathbf{C}_t),
\]
where $\mathbf{Z}_t=f_{\mathrm{enc}}(\mathbf{X}_t,\mathbf{M}_t)$ is the scene latent and $T_f$ is the rollout horizon. Although the architecture predicts controls from $t$ to $t+T_f-1$, only the first-step control
$\hat{\mathbf{U}}_t$ is executed. The scene history is then updated, and the full pipeline is queried again at the next step.

This receding-horizon procedure is analogous to model predictive control: the controller selects the current action using a finite-horizon prediction that accounts for future interaction consequences, while continuously re-planning as the scene evolves. This allows the system to maintain long-horizon awareness without committing to long open-loop rollouts that are prone to drift.

\subsection{Hybrid Co-Training for Policy Stabilization}

\begin{table*}[ht]
\centering
\caption{Quantitative comparison of closed-loop simulation results. Hard acceleration and sharp turn incidents are defined by thresholds exceeding $2.5 \text{ m/s}^2$ and $20^\circ\text{/s}$, respectively. Safety flags indicate critical scenarios where the Time-to-Collision (TTC) falls below 1.5 seconds.}
\renewcommand{\arraystretch}{1.5} 
\arrayrulecolor{white} 
\begin{tabular}{|l|c|c|c|c|}
\hline
\rowcolor{headerblue} 
\cellcolor{headerblue} & \color{white}\textbf{STEER under Gigaflow Architecture} & \color{white}\textbf{Our Method} & \color{white}\textbf{Wayformer} & \color{white}\textbf{MotionLM}
 \\ \hline

\rowcolor{rowgrey} Average speed (m/s) & 5.45 & 5.45 & 5.46 & 5.84 \\ \hline
\rowcolor{rowgrey} Hard Accel / km & 0.312 & 0.169 & 0.169 & 0.168 \\ \hline
\rowcolor{rowgrey} Sharp Turn / km & 5.0e-2 & 4.72e-2 & 4.80e-2 & 5.61e-2 \\ \hline
\rowcolor{rowgrey} Safety Flag / km & 7.58e-3 & 4.65e-3 & 1.08e-2 & 1.27e-2 \\ \hline
\rowcolor{rowgrey} Collision / km & 3.28e-3 & 2.37e-3 & 4.82e-3 & 6.19e-3 \\ \hline
\rowcolor{rowgrey} ADE loss & N/A & 0.527 & 0.612 & 1.762 \\ \hline
\end{tabular}
\label{table1}
\end{table*}

A key challenge in closed-loop deployment is distribution shift induced by compounding execution errors. Once the generated controls deviate from the training distribution, future scene states may drift into out-of-distribution regions, which can further degrade both decision quality and control stability. To mitigate this effect, we employ a hybrid co-training scheme that combines high-level MARL optimization with auxiliary supervision for the low-level motion module.

Concretely, we introduce a heuristic expert that provides corrective recovery trajectories or controls when drift occurs. In our current setup, the expert acts as a recovery mechanism that periodically realigns the agent toward feasible high-reward behavior. Let $\mathbf{Y}_t^{\mathrm{exp}}$ denote the expert recovery target associated with the current scene. The low-level module is trained to follow this recovery signal through an auxiliary reconstruction objective:
\[
\mathcal{L}_{\mathrm{traj}} = \left\|\hat{\mathbf{Y}}_t - \mathbf{Y}_t^{\mathrm{exp}}\right\|_2^2,
\]
where $\hat{\mathbf{Y}}_t$ denotes the predicted spatial trajectory rollout induced by executing the command-conditioned low-level controls.

To encourage physically reasonable execution, we further regularize the low-level output with smoothness and safety terms:
\[
\mathcal{L}_{\mathrm{low}}
=
\mathcal{L}_{\mathrm{traj}}
+ \lambda_s \mathcal{L}_{\mathrm{smooth}}
+ \lambda_c \mathcal{L}_{\mathrm{coll}},
\]
where $\mathcal{L}_{\mathrm{smooth}}$ penalizes abrupt control variation and $\mathcal{L}_{\mathrm{coll}}$ penalizes collision-prone behavior. These regularizers help the learned controller remain stable and safe even when strategic commands induce aggressive or uncommon scene transitions.

Meanwhile, the high-level policy is optimized with MARL rewards designed to encourage long-horizon coordination and task completion. Denoting the high-level objective by
\[
J_{\mathrm{high}} = \mathbb{E}_{\pi_{\mathrm{high}}}\Bigl[\sum_{t=0}^{T}\gamma^t r_t\Bigr],
\]
the overall training alternates between strategic policy improvement at the high level and supervised stabilization at the low level. Intuitively, the high-level policy explores proactive interaction strategies that are difficult to obtain from imitation alone, while the low-level module is kept grounded by recovery supervision and physical regularization.

This hybrid co-training scheme is important for two reasons. First, it reduces the sensitivity of the command-conditioned low-level model to out-of-distribution rollout states. Second, it prevents the full hierarchical system from collapsing into unstable behavior during joint training, thereby making the closed-loop policy substantially more robust than naive end-to-end optimization.

\subsection{Summary of the Hierarchical Pipeline}
The overall pipeline is summarized below and illustrated in Fig.~\ref{fig: 31}. Starting from the current scene history and context, the high-level Stackelberg-style MARL module generates strategic commands sequentially over agents:
\[
\mathbf{C}_t = \pi_{\mathrm{high}}(s_t).
\]
The low-level command-conditioned Wayformer then combines these commands with the scene latent to produce a finite-horizon control rollout:
\[
\hat{\mathbf{U}}_{t:t+T_f-1}
=
f_{\mathrm{dec}}\!\bigl(f_{\mathrm{enc}}(\mathbf{X}_t,\mathbf{M}_t),\mathbf{C}_t\bigr).
\]
Finally, only the first-step control $\hat{\mathbf{U}}_t$ is applied in a sliding-window closed-loop manner, and hybrid co-training is used to stabilize learning under distribution shift. This decomposition preserves explicit interaction reasoning at the strategic level while retaining continuous low-level motion realization for dense multi-agent control.

\section{EXPERIMENT}
 \begin{figure}[t]
    \centering       \includegraphics[width=.8\linewidth]{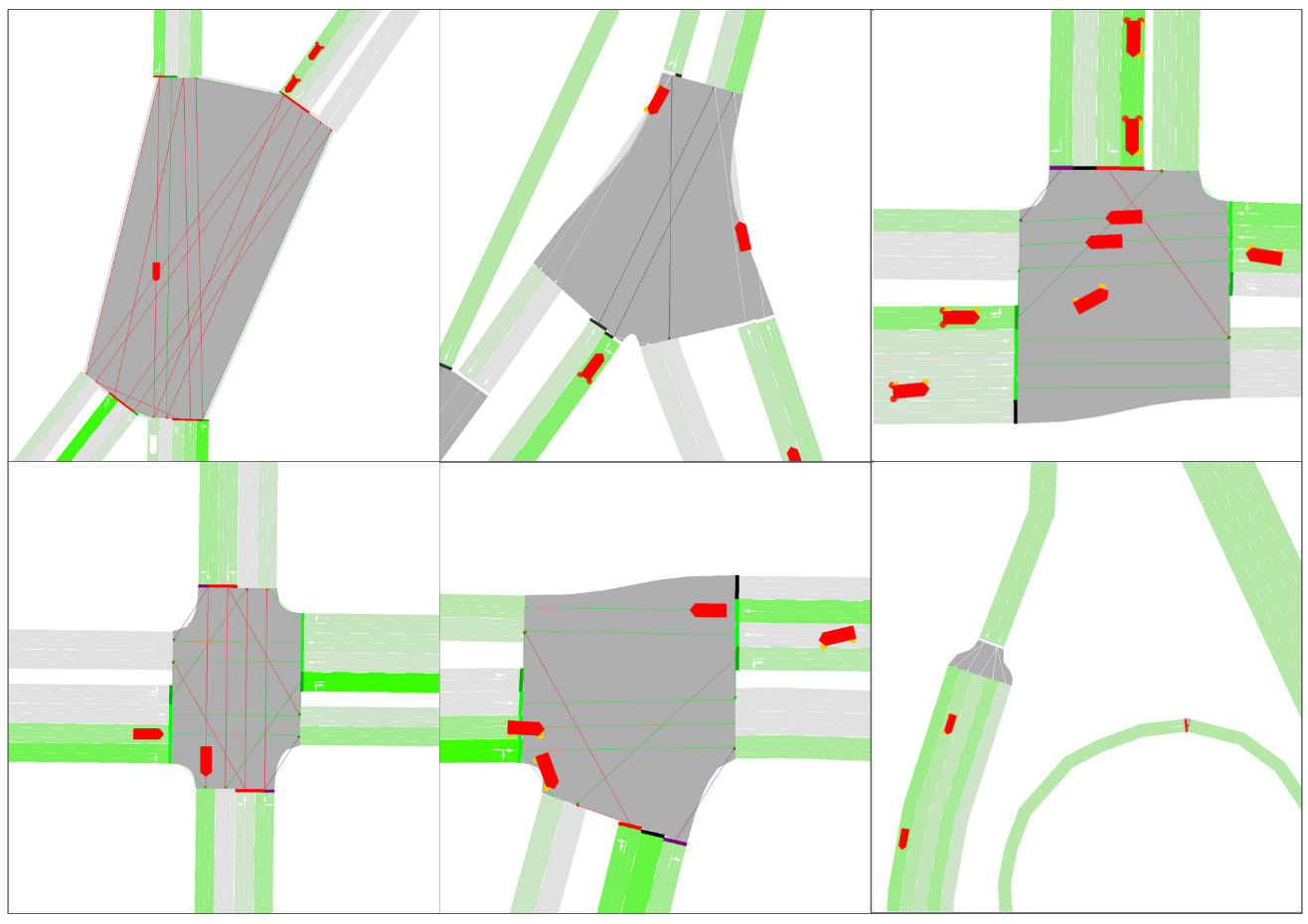}
    \caption{SUMO-based testing network covering a $1.5 \times 2$ mile urban region in California.}
   \label{fig: world}
 \end{figure}

\begin{figure*}[t]
\centering   
 \begin{subfigure}{\textwidth}
    \centering
    \includegraphics[width=.8\linewidth]{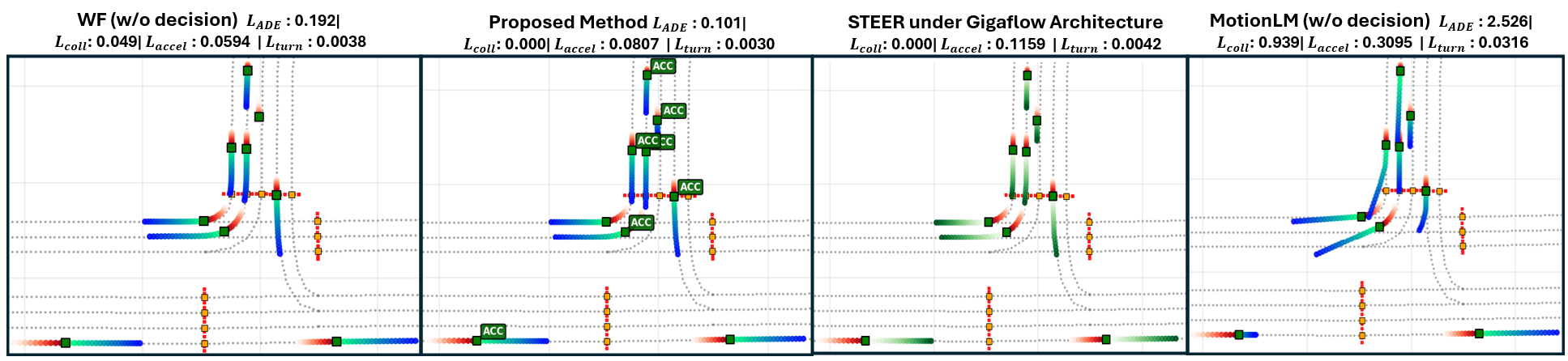}
  \caption{In complex dense scenarios, the proposed method demonstrates lower ADE and collision loss compared to the strategically uninformed realization module, alongside reduced penalties for hard braking and sharp turns.}
   \label{fig: c_no_yield}
 \end{subfigure}%

\medskip

 \begin{subfigure}{\textwidth}
     \centering
    \includegraphics[width=.8\linewidth]{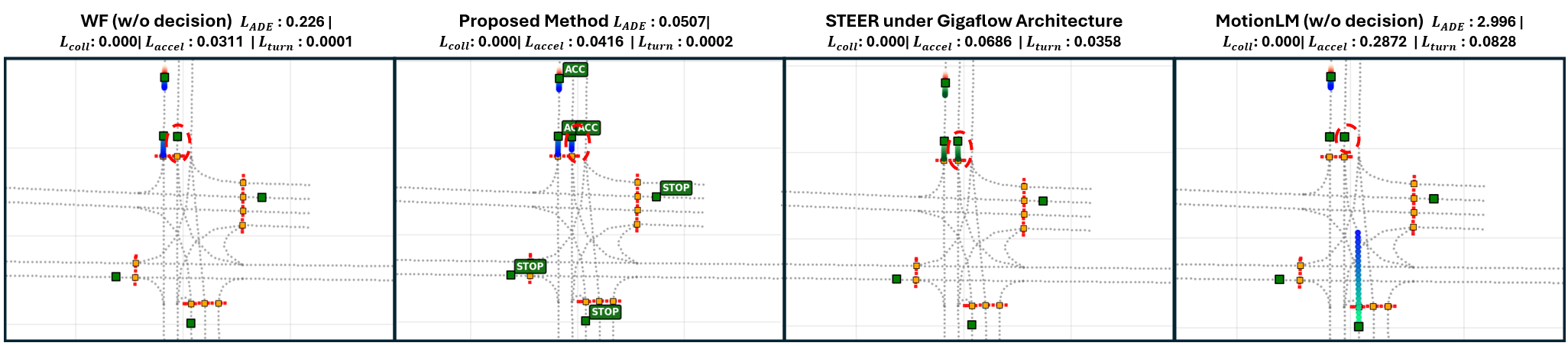}
    \caption{The proposed method highlights the importance of decision integration, enabling the vehicle to accelerate from a stop while the strategically uninformed realization module  baseline remains idle (indicated by the red dashed circles).}
    \label{fig: c_yield}
 \end{subfigure}
  \caption{Illustrative comparisons of multi-vehicle control using strategically uninformed realization module (Wayformer), the proposed method, and Gigaflow self-play. Performance is evaluated across average displacement error ($L_{ade}$), collision loss ($L_{coll}$), acceleration penalty ($L_{accel}$), and sharp turn penalty ($L_{turn}$). In both scenarios, the proposed method achieves lower ADE and collision loss compared to the Gigaflow baseline, while maintaining the advantage of strategic awareness.}
  \label{fig: 43}
 \end{figure*}

\subsection{Experimental Setup}

We implement our framework using the SUMO traffic simulator, leveraging its scalable environment for multi-agent interaction. Our interface enables the execution of low-level, continuous control actions:  acceleration and steering angle. By setting the simulation step length to $0.1$s, we approximate continuous vehicle dynamics while satisfying kinematic constraints.

We evaluate our approach in a simulation spanning a $1.5 \times 2$ mile urban network in California, as shown in Fig. \ref{fig: world}. The network comprises $9$ topologically diverse junctions—including multi-way intersections, freeway ramps, and dedicated turn lanes—each parameterized by precise turning ratios, lane semantics, and signal schedules. This environment contains nearly 300 vehicles simultaneously. Ultimately, this setup provides a comprehensive benchmark for testing closed-loop, multi-agent traffic simulation under dense, geometrically complex, and long-horizon traffic conditions.

\subsection{Quantitative Evaluation and Ablation Analysis}
We evaluate the proposed hierarchical architecture against baselines that serve as structured ablations. STEER under Gigaflow utilizes rule-based execution, ablating our learned low-level continuous realization. Conversely, the standalone Wayformer and MotionLM act as passive predictors, representing conventional imitation learning agents and thereby ablating our high-level interaction reasoning module. The results are summarized in Table~\ref{table1}.

\textit{1) Kinematic Smoothness:} Integrating a spatial-temporal world model for trajectory synthesis fundamentally improves motion quality. While our method matches the operational efficiency of the rule-based STEER baseline, it significantly reduces harsh longitudinal accelerations and sharp lateral maneuvers. This confirms that replacing rigid heuristic execution with learned continuous realization effectively dampens abrupt control inputs and mitigates the fragility of rule-based controllers.

\textit{2) Closed-Loop Safety:} The safety metrics strongly validate the need for explicitly decoupling strategic intent from trajectory generation. While end-to-end predictors like MotionLM yield marginally higher average speeds, they suffer from notably higher collision and safety flag rates due to a lack of proactive coordination. Similarly, the standalone Wayformer struggles with closed-loop safety in dense multi-agent interactions. By utilizing reinforcement learning for high-level action selection, our architecture effectively anticipates complex interactions, achieving the lowest collision rates and fewest safety interventions across all evaluated methods.

\textit{3) Trajectory Precision:} Beyond safety and smoothness, our architecture demonstrates superior spatial accuracy. When evaluated on average displacement error (ADE), our method outperforms the standalone predictive baselines. This indicates that conditioning the low-level continuous realization on explicit high-level strategic intent provides critical contextual grounding, allowing the model to generate spatial trajectories that more accurately reflect optimal multi-agent coordination.

\textit{4) Architectural Synergy:} These comparisons demonstrate that removing high-level strategic reasoning critically weakens coordinated decision-making, while relying on heuristic execution degrades kinematic smoothness. The superior performance of our framework confirms that safe, naturalistic navigation requires the active synthesis of both explicit multi-agent reasoning and robust spatial-temporal execution.

\subsection{Qualitative Analysis}

To further evaluate our architecture, we visualize predicted trajectories in representative intersection scenarios (Fig.~\ref{fig: 43}) to compare closed-loop performance against the Wayformer and Gigaflow (self-play) baselines.

The proposed method generates spatially consistent, interaction-aware trajectories, achieving lower Average Displacement Error ($L_{ade}$) and collision loss ($L_{coll}$) in dense traffic. Agents exhibit coordinated behaviors, such as timely acceleration and safe distance maintenance, which stems from integrating high-level strategic reasoning with continuous low-level trajectory realization.

In contrast, baselines struggle with coherent long-horizon interactions. Passive predictors like Wayformer lack explicit strategic integration, limiting their ability to make informed decisions in complex closed-loop settings. As highlighted by the red dashed circles in Fig.~\ref{fig: 43}(b), this weak strategic awareness causes Wayformer to remain passively idle instead of proactively accelerating from a stop. Conversely, architectures lacking learned low-level realization, such as Gigaflow (self-play), produce abrupt or rigidly constrained motions, reflected in higher acceleration ($L_{accel}$) and sharp turn ($L_{turn}$) penalties.

The proposed architecture avoids Wayformer's strategically uninformed idling and Gigaflow's abrupt kinematics, yielding smooth and decisive driving. These qualitative results align with our quantitative metrics, underscoring the benefits of explicit decision integration and continuous trajectory realization.

\section{DISCUSSION}
A key contribution of this work lies in the architectural decomposition itself. By separating high-level interaction reasoning from low-level continuous trajectory realization, the framework provides a structured interface through which strategic decisions can influence scene evolution without requiring the decision module to solve the full continuous control problem directly. While the architecture is agnostic to the choice of imitation backbone, our co-training experiments with the Stackelberg Decision Transformer demonstrate that continuous-space inductive biases are necessary for self-play-based architectures to move beyond collision-avoidance equilibria toward long-horizon, interaction-aware behavior fidelity

This design offers two practical benefits. First, it improves modularity and extensibility: the high-level module can be instantiated with different multi-agent decision-making algorithms, while the low-level module can be implemented using different state-of-the-art motion predictors or controllers. Second, it increases the flexibility of discrete strategic decisions by realizing them through continuous trajectory generation rather than fixed motion primitives. As a result, the architecture preserves the tractability of structured high-level decision-making while avoiding the rigidity of purely discrete trajectory execution.


\section{CONCLUSION}
In this paper, we introduced a hierarchical framework for resolving the trade-off between behavioral realism and scalability in closed-loop traffic simulation. By decoupling multi-agent interaction reasoning from continuous trajectory realization, our approach bridges the gap between strategic intent and physically consistent motion realization. We integrated a Stackelberg-style MARL decision module with a low-level continuous trajectory realization module, and employed a hybrid co-training scheme to mitigate distribution shift during closed-loop execution. Evaluations in dense urban environments simulated in SUMO demonstrate that our method significantly outperforms predictive and rule-based baselines, achieving superior kinematic smoothness, spatial accuracy, and collision reduction. Ultimately, this architecture provides a modular foundation for scalable multi-agent coordination. Future work will focus on accommodating broader agent heterogeneity, grounding the high-level module with naturalistic driving data to better capture socially aware human behavior, and introducing safety-critical behavior modeling to capture long-horizon edge cases within the hierarchical framework.


\bibliographystyle{IEEEtran}
\bibliography{Ref/Intro}

\end{document}